\documentclass{article}

\PassOptionsToPackage{numbers, compress}{natbib}
\usepackage[preprint]{neurips_2025}

\usepackage[utf8]{inputenc} 
\usepackage[T1]{fontenc}    
\usepackage{hyperref}       
\usepackage{url}            
\usepackage{booktabs}       
\usepackage{amsfonts}       
\usepackage{nicefrac}       
\usepackage{microtype}      
\usepackage{xcolor}         
\usepackage{graphicx}
\usepackage{wrapfig}
\usepackage{subcaption}
\usepackage{longtable}
\usepackage{pifont}
\usepackage{multirow}

\usepackage{blindtext}
\usepackage{tcolorbox}
\usepackage{amsmath}
\usepackage{soul}

\newcommand{{\FairPP}}{\textsc{Fair-PP}}
\definecolor{left}{RGB}{70, 152, 198}
\definecolor{right}{RGB}{221, 108, 90}

\title{\FairPP: A Synthetic Dataset for Aligning LLM with Personalized Preferences of Social Equity}

\author{%
Qi Zhou$^{1}$
\quad Jie Zhang$^{2}$ \quad Dongxia Wang$^{1}$\thanks{Corresponding author}  \quad
\textbf{Qiang Liu}$^1$  \\ \textbf{Tianlin Li}$^{3}$ \quad \textbf{Jin Song Dong}$^{4}$ \quad 
\textbf{Wenhai Wang}$^{1}$  \quad \textbf{Qing Guo}$^{2}$ \\
$^1$Zhejiang University \quad $^2$IHPC and CFAR, A*STAR \quad $^3$Nanyang Technological University \\ $^4$National University of Singapore
\\
\{qi.zhou, qiang.liu, dxwang, zdzzlab\}@zju.edu.cn,
tianlin001@e.ntu.edu.sg,\\
zhang\_jie@cfar.a-star.edu.sg,
tsingqguo@ieee.org,
dcsdjs@nus.edu.sg
}
\begin{document}

\maketitle

\begin{abstract}
Human preference plays a crucial role in the refinement of large language models (LLMs). However, collecting human preference feedback is costly and most existing datasets neglect the correlation between personalization and preferences. To address this issue, we introduce {\FairPP}, a synthetic dataset of personalized preferences targeting social equity, derived from real-world social survey data, which includes 28 social groups, 98 equity topics, and 5 personal preference dimensions. Leveraging GPT-4o-mini, we engage in role-playing based on seven representative persona portrayals guided by existing social survey data, yielding a total of 238,623 preference records. Through {\FairPP}, we also contribute (i) An automated framework for generating preference data, along with a more fine-grained dataset of personalized preferences; (ii) analysis of the positioning of the existing mainstream LLMs across five major global regions within the personalized preference space; and (iii) a sample reweighting method for personalized preference alignment, enabling alignment with a target persona while maximizing the divergence from other personas. Empirical experiments show our method outperforms the baselines.

\raisebox{-0.3\height}{\hspace{0.05cm}\includegraphics[width=0.45cm]{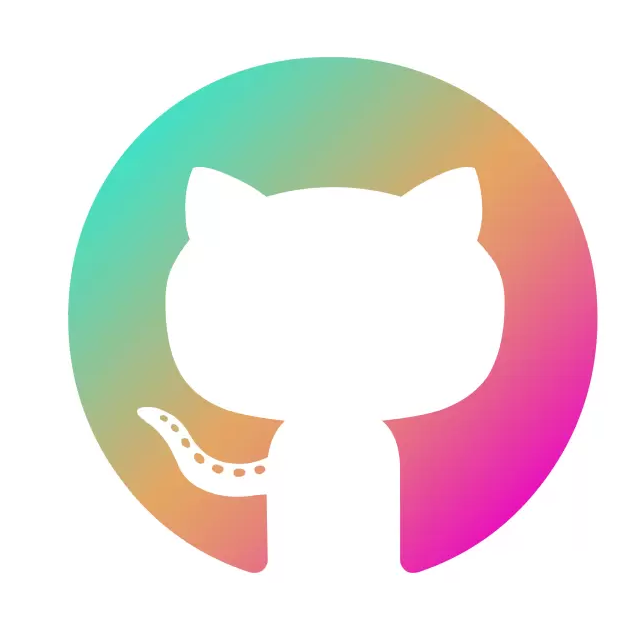}} \small \textbf{\mbox{Code:}} \href{https://github.com/tools-only/FairPP}{github.com/tools-only/FairPP} \\
\vspace{1em}
\raisebox{-0.3\height}{\includegraphics[width=0.4cm]{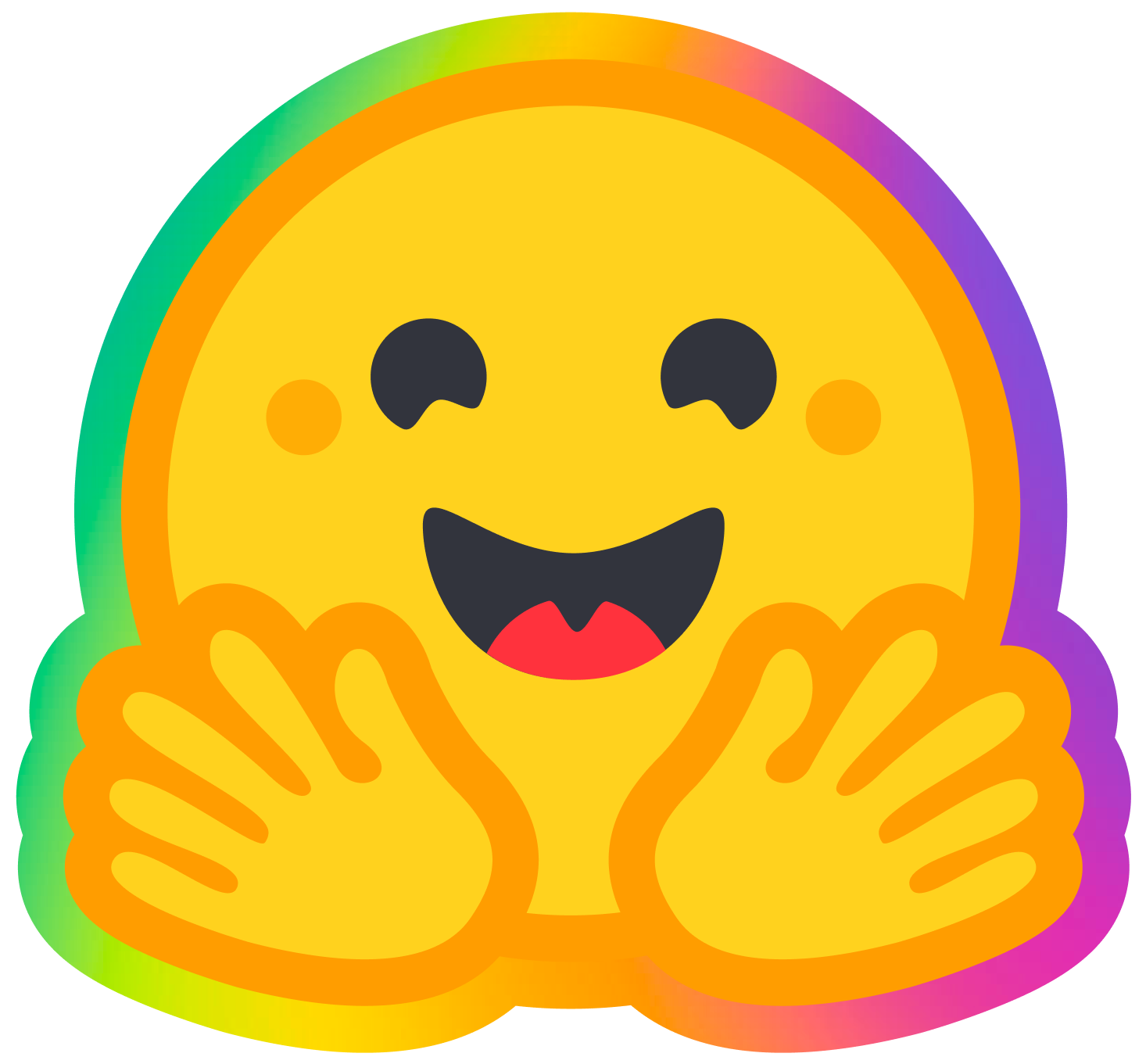}} \small \textbf{\mbox{Dataset:}} \href{https://huggingface.co/datasets/tools-o/Fair-PP}{huggingface.co/datasets/tools-o/Fair-PP}

\end{abstract}


\section{Introduction}

With the growing adoption of LLMs in public policy making and public services\footnote{\url{https://openai.com/global-affairs/introducing-chatgpt-gov/}}\footnote{\url{https://openai.com/index/democratic-inputs-to-ai-grant-program-update/}}\footnote{\url{https://huggingface.co/stewhsource/GovernmentGPT}}, a key question is \emph{How can an LLM-based public policy maker accurately capture and represent dynamic and diverse personalized preferences of social values?} On one hand, social values like equity vary significantly among individuals\cite{huseman1987new, tuli2023understanding}.
On the other hand, to promote the alignment of LLMs with diverse societal values is crucial to achieve societal safety \cite{ji2023beavertails, qi2024safety, yin2024safeworld, huang2024lisa}, supporting cultural inclusivity~\cite{tao2024cultural, alkhamissi2024investigating, li2024culturellm, li2024culturepark}, and also reflecting diverse human values \cite{durmus2023towards, santurkar2023whose, sorensen2024value, zhao2024worldvaluesbench}. 
For the alignment, high-quality human feedback of their social-value preference is necessary.
However, its collection can incur significant costs \cite{dubois2023alpacafarm, cui2023ultrafeedback}.
Moreover, human personalized preferences within a region evolve over time, influenced by factors such as demographic shifts and social development \cite{greenfield2016social, ramos2019humans, zarwi2017modeling}.

\begin{wrapfigure}{r}{0.40\textwidth} 
    \centering
    \includegraphics[width=0.4\textwidth]{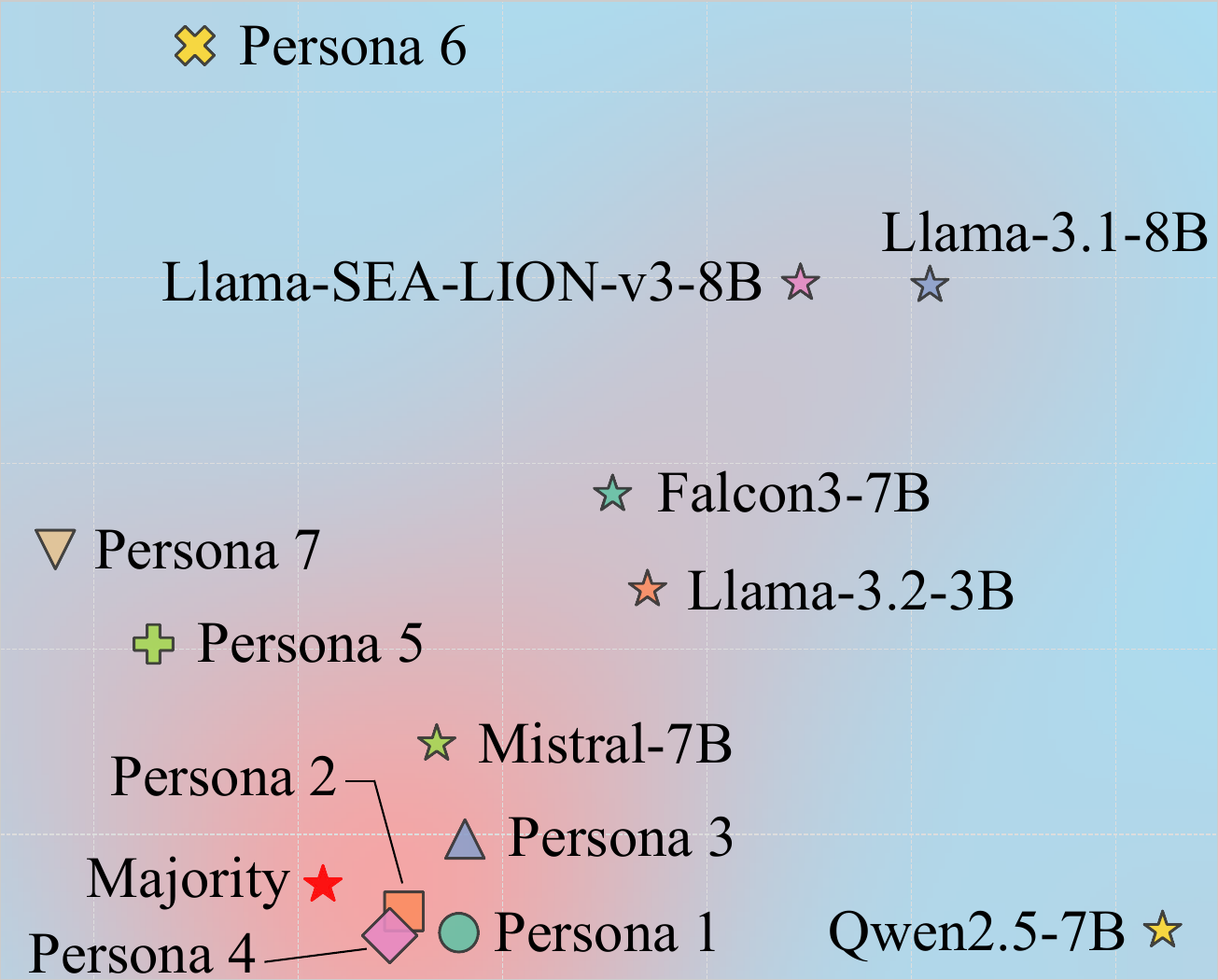}
    \vspace{-0.6cm}
    \caption{Landscape of {\FairPP} space.}
    \label{fig:space}
    \vspace{-0.4cm}
\end{wrapfigure}
 Despite increasing efforts in exploring and collecting human preference data of social values, existing datasets face several limitations: (i)  \textbf{Social equity unaddressed}: lack consideration of social equity which is an important issue in social psychology\footnote{\url{https://www.un.org/en/actnow/ten-actions-just-society}}
(ii) \textbf{Neglecting personalization}: Most of them emphasize universal viewpoints, \emph{e.g.,} cultural and political beliefs \cite{li2024culturepark}\, lacking consideration of the relation between preference and personalization \cite{huseman1987new, king1994measurement}. 
(iii) \textbf{Single-dimensional}: with data coverage insufficient to capture personalized preferences across diverse dimensions. (iv) \textbf{Limited information resources}: they generate data based on limited information sources, typically those from the computer science community or manually hand-crafted, failing to reflect real-world diversity. 
Last but not the least, typically alignment techniques are emphasized with less focus on dataset design and analysis. 

\begin{figure}[h]
    \centering
    \includegraphics[width=1\textwidth]{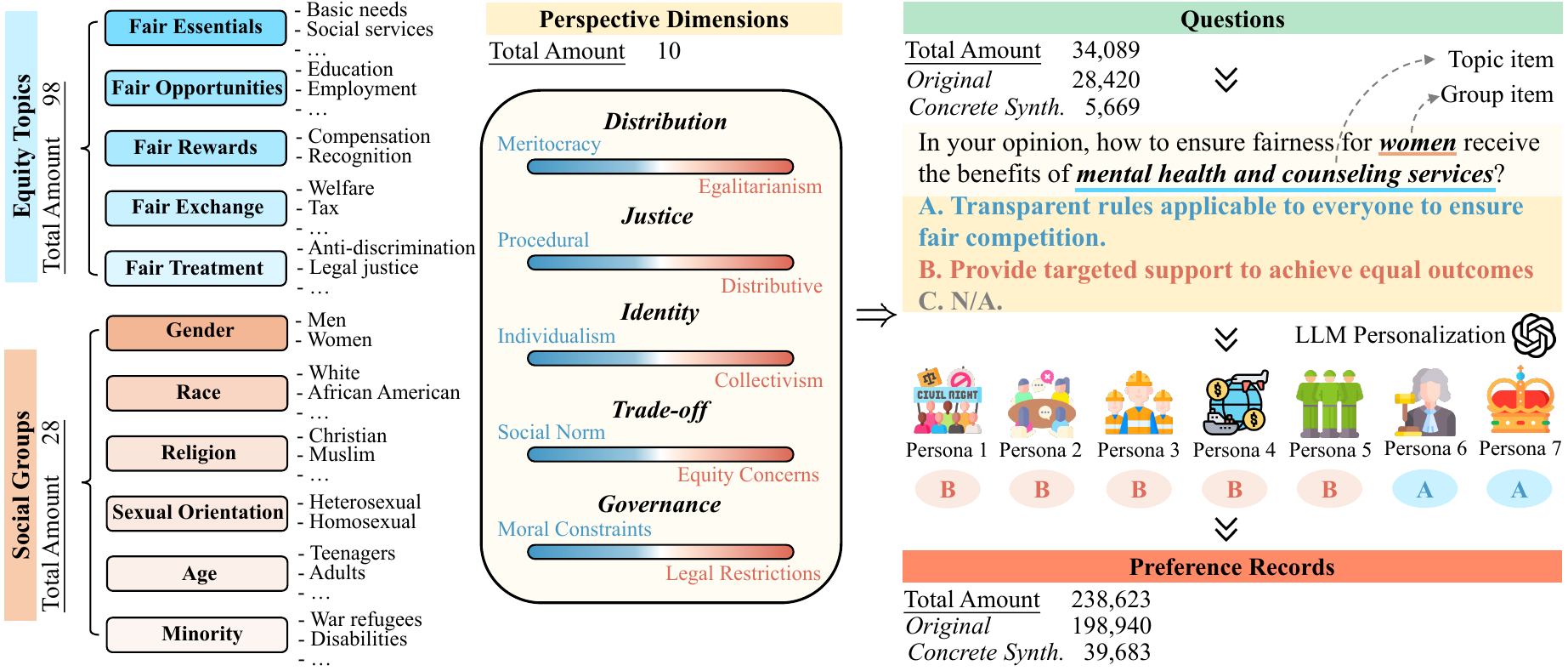} 
    \caption{An overview of the {\FairPP} dataset. Each question consists of three parts: the social group, an equity topic, and a perspective dimension. An example question is shown on the right where option A and B represent two types of viewpoints under a specific dimension. Personalized preferences are collected through LLM personalization that leverages 7 value portrait based on the real-world social surveys.}
    \label{fig:2}
    \vspace{-0.3cm}
\end{figure}

We release {\FairPP}, a new personalized preference resource capturing fine-grained preferences of social equity in practical contexts, enabling exploration of mainstream LLMs’ preference landscapes across global regions (as shown in Figure \ref{fig:space}) and supporting further preference alignment. We develop a comprehensive question bank based on the integration of three key components: \ding{182} a collection of 28 common social groups, \ding{183} a collection of 98 practical topics derived from fine-grained expansion of five categories of fair necessities\footnote{\url{https://fairnessfoundation.com/fairnecessities}\label{fairnecessities}}, as defined by The Fairness Foundation, a social research charity focused on social equity, and \ding{184} five diverse perspective dimensions based on core equity concepts, each encompassing two contrasting viewpoints. Furthermore, based on a subset of these questions, more concrete and realistic scenario simulation data were generated as a supplement for testing. Following the seven value‑based segments of the UK public as defined in ``More in Common'' and YouGov’s 2020 Britain’s Choice report \cite{surridge2021britain}, we utilize GPT-4o-mini for LLM personalization to collect personalized preference data across our questionnaire. As shown in Figure \ref{fig:2}, {\FairPP} contains 34,089 survey questions and 238,623 personalized preference data points, with key features including \textbf{sources from real social survey}, \textbf{comprehensive content}, and \textbf{automated data generation}. 

To the best of our knowledge, {\FairPP} is the first personalized preference dataset targeting social equity values. By anchoring the seven distinct personas with diverse personalized preferences, {\FairPP} maps out a personalized preference space, which paves the way for exploring the position of current mainstream LLMs across various global regions (Section \ref{sec:case_1}). In addition, {\FairPP} enables personalized preference alignment, we propose a sample-level weighting mechanism to enhance the distinctiveness of target personas. Experiments on two test sets show that our methods successfully align with the target persona while maximizing the divergence from others (Section \ref{sec:case_2} and \ref{sec:case_3}). Our contributions can be summarized as follows:
\begin{enumerate}
    \item We introduce {\FairPP}, a dataset targeting personalized equity preferences, derived from real-world surveys and covering multiple social groups, equity topics, value dimensions, and representative personas.
    \item We present a framework for automatically generating preference data, leveraging questionnaire data from {\FairPP}, where LLMs can autonomously generate large-scale preference data. This serves as a valuable complement to real preference data, while also effectively adapting to dynamic and diverse equity contexts
    \item We demonstrate the value of {\FairPP} by positioning the mainstream LLMs in the personalized equity preference space using preference anchor points, as well as targeted preference alignment. Additionally, we propose a sample-weighted preference alignment enhancement method, which effectively improves the alignment performance.
\end{enumerate}

\section{{\FairPP}}
\label{dataset}
{\FairPP} is a multi-level resource for personalized equity preferences, features questions structured around three key components: social groups, equity topics, and perspective dimensions, which are further detailed in Section \ref{sec:3.1}--\ref{sec:3.3}. The methodology for question generation is subsequently introduced in Section \ref{sec:3.4}. Subsequently, we present the personalized preferences captured through LLM-personalization in Section \ref{sec:3.5}.

\subsection{Social Groups} 
\label{sec:3.1}
{\FairPP} covers a total of 28 social groups, including common standard social group categories like gender, age, race, religion, and sexual orientation, as well as a range of typical social minority groups such as 911 victims, Black Lives Matter supporters, war refugees, famine victims, feminists, and environmentalists. These selections reflect representative social concerns and historical contexts from various regions worldwide. More details are presented in Appendix \ref{appendix:groups_topics}. 

\subsection{Social Equity Topics}
\label{sec:3.2}
To integrate real-world equity preferences during dataset construction, {\FairPP} draws upon five major equity topics identified through surveys conducted by The Fairness Foundation, a real-world social research organization. Building on these, we consider a comprehensive set of subtopics including basic living needs, healthcare, education, employment, finance, law, and other relevant social issues. An overview of these topics is provided in Figure \ref{fig:2}, with the specific categories listed as follows:

\textbf{Fair Essentials}. Meeting people's basic needs is fundamental to achieving social equity. Within the concept of fair essentials, we identify four fundamental needs:
(1) Basic material needs: this encompasses the essentials for survival and well-being, such as food, clean water, and shelter.
(2) Basic health needs: access to essential medications, basic sanitation, and healthcare services are crucial for maintaining health.
(3) Basic social services: everyone deserves to feel safe, have access to public transportation, and receive the basic education, enabling them to participate fully in society.
(4) Fundamental rights: human rights, freedom of speech, and other fundamental freedoms are essential for individual autonomy and dignity.

\textbf{Fair Opportunities}. Everyone deserves the chance to achieve success in life. We categorize fair opportunities into three key areas: (1) Education and skills development: access to affordable higher education, vocational training, and lifelong learning opportunities empowers individuals to gain the knowledge and skills needed to thrive. (2) Economic and employment: this encompasses fair access to jobs, opportunities for advancement, and career switch, ensuring that everyone has the chance to contribute to the economy and achieve economic security. (3) Political participation: including exercising right to vote, running for office, and policy feedback access, which offers avenues for citizens to engage in public policy.

\textbf{Fair Rewards}. This principle emphasizes that everyone should be justly rewarded for their efforts and contributions. To explore this concept, we identify two main categories: (1) Compensation: including wages, bonus and tips, which focuses on physical rewards within the workplace. (2) Social recognition: recognizing and appreciating individual efforts publicly, such as verbal praise and media shout-outs.

\textbf{Fair Exchange}. Aims to ensure a balance between individuals' social welfare and tax payments, which can be broadly categorized into three main areas: (1) Reciprocity: focuses on providing support and benefits to individuals, such as unemployment benefits and disability supports. (2) Welfare: encompasses a range of services and programs designed to improve the well-being of individuals and families, including subsidized childcare, free legal aid, and mental health counseling. (3) Tax: various taxes levied on individuals and businesses to fund social welfare programs and public services, typically with income tax, inheritance tax and luxury tax.

\textbf{Fair Treatment}. Fair Treatment ensures that people are treated equitably in all aspects of society. For this topic, we categorize three key themes: (1) Anti-discrimination: this includes protection against stigmatization, culturally inclusive healthcare services, and more. (2) Legal and social justice: This encompasses protection from workplace harassment, safeguards against exploitative contracts, and other measures. (3) Public resource equity: which involves initiatives such as the distribution of public restrooms in underserved areas, and support for public housing programs.

For each subtopic, we further divide it into more specific subject matters, eventually resulting in a total of 97 specific topics. For details, please refer to Appendix \ref{appendix:groups_topics}.

\subsection{Personalized Preference Dimensions}
\label{sec:3.3}
For each group and topic, we design five personalized preference dimensions informed by real-world social viewpoints, each including two distinct orientations. Specifically,

\begingroup
\addtolength\leftmargini{-25pt}
\begin{quote}
\vspace{-0.2cm}
     Dimension 1 \textbf{({Meritocracy}} \textit{vs.} \textbf{{Egalitarianism})} \cite{goto2022belief}:  \emph{Should we prioritize \textbf{\underline{\textcolor{left}{current achievements}}} or promoting \textbf{\underline{\textcolor{right}{evenly outcomes}}}?}
\vspace{-0.2cm}
\end{quote}
This dimension contrasts two approaches to equity: meritocracy, where rewards are based on achievements (\emph{e.g.}, promotions based on performance), and egalitarianism, which emphasizes evenly outcomes (\emph{e.g.}, distributing resources evenly). The debate centers on whether merit or equality should be prioritized in equity judgments.
\begin{quote}
\vspace{-0.2cm}
     Dimension 2 \textbf{({Procedural}} \textit{vs.} \textbf{{Distributive})} \cite{clay2005procedural}: \textit{Should justice emphasize \textbf{\underline{\textcolor{left}{fair competitive}}} or prioritize \textbf{\underline{\textcolor{right}{supporting the disadvantaged}}} to achieve equal outcomes?}
\vspace{-0.2cm}
\end{quote}
Dimension 2 highlights the contrast between two conceptions of equity: procedural justice (fair processes, e.g., decisions based on neutral rules like standardized testing) and distributive justice (fair outcomes, e.g., corrective policies like affirmative action). The core issue is whether equity depends on impartial procedures or equitable results.
\begin{quote}
\vspace{-0.2cm}
     Dimension 3 \textbf{({Individualism}} \textit{vs.} \textbf{{Collectivism})} \cite{lefebvre2013culture}: \textit{Should resources be shared based on \textbf{\underline{\textcolor{left}{individual efforts}}} or \textbf{\underline{\textcolor{right}{collective allocation}}}?}
\vspace{-0.2cm}
\end{quote}
This dimension addresses whether equity should emphasize individual responsibility and effort or prioritize collective well-being by emphasizing social responsibility, highlighting differing perspectives on how equity is understood either through personal contribution or through shared obligations and group oriented outcomes.
\begin{quote}
\vspace{-0.2cm}
     Dimension 4 \textbf{({Social norm}} \textit{vs.} \textbf{{Equity concerns})} \cite{bussolo2024social}: \textit{Should we prioritize adherence to established \textbf{\underline{\textcolor{left}{social norms}}} or \textbf{\underline{\textcolor{right}{the pursuit of equity}}}?}
\vspace{-0.2cm}
\end{quote}
This dimension contrasts social norms (\emph{e.g.}, maintaining traditional gender roles) with equity concerns (\emph{e.g.}, advocating for gender equality in the workplace). The question is whether to preserve tradition or to promote equity, even if it challenges societal conventions.
\begin{quote}
\vspace{-0.2cm}
     Dimension 5 \textbf{({Moral}} \textit{vs.} \textbf{{Law})} \cite{alder2006achieving}: \textit{Should equity be achieved primarily through \textbf{\underline{\textcolor{left}{moral constraints}}} or \textbf{\underline{\textcolor{right}{legal constraints}}}?}
\vspace{-0.2cm}
\end{quote}
This dimension examines whether equity should be guided primarily by moral principles or by legal constraints. It addresses the question of whether ethical considerations or formal legal frameworks ought to serve as the foundation for fair treatment.
\endgroup
\vspace{-0.2cm}

\subsection{Question Data Generation}
\label{sec:3.4}
\textbf{Template-based Data.} We first create a multiple-choice questionnaire with a total of 28,420 samples, where questions combined social groups, equity topics, and perspective dimensions, as described in the sections above. Corresponding to its specific perspective dimension category, each question included three options: two opposing viewpoints and an N/A option to avoid bias due to forced selections.\looseness=-1

\textbf{Generation-based Data.} Furthermore, to improve the diversity of the data, we sample 5,669 questions from the dataset and generate concrete scenario samples using GPT-4o. For each question and its corresponding options, we prompt the model to create short real-world scenarios that reflect each perspective. Each scenario includes the background of a fictional person, a decision point where they receive a service or resource, and a brief emotional response. This variant process yields more realistic and diverse scenarios, enabling the in-depth analysis of personalized preferences across a variety of situations. For more details, please refer to Appendix \ref{appendix:data}, which includes the detailed prompts and an example data point presented in Figure \ref{fig:data_point}.
\subsection{Personalized Preference}
\label{sec:3.5}
\textbf{Persona} Drawing on the social persona typologies derived from real-world surveys conducted by More in Common, which identify seven personas in UK society\footnote{\url{https://www.britainschoice.uk/segments/}}, we anonymize the country-related descriptions (\emph{e.g.,} preferred media outlets and supported political parties) to construct universal persona profiles.
Specifically, the key summary for these seven personas is as follows:
\begingroup
\addtolength\leftmargini{-25pt}
\begin{quote}
\textbf{Persona 1} (\textit{Progressive Activists}): Advocate for social justice, equality, and systemic change.

\textbf{Persona 2} (\textit{Civic Pragmatists}): Emphasize civic responsibility, community cohesion, and pragmatic compromise.

\textbf{Persona 3} (\textit{Disengaged Battlers}): Focused on day-to-day survival, often feeling excluded from broader political discourse.

\textbf{Persona 4} (\textit{Established Liberals}): Comfortable with diversity and change; confident in their social and cultural identity.

\textbf{Persona 5} (\textit{Loyal Nationals}): Proud of their country’s heritage and history; emphasize patriotism and national identity.

\textbf{Persona 6} (\textit{Disengaged Traditionalists}): Value self-reliance, routine, and hard work; often skeptical of rapid change.

\textbf{Persona 7} (\textit{Backbone Conservatives}): Critical of immigration and identity politics; support reduced government spending.  
\end{quote}
\endgroup

For more detailed persona prompts, please refer to Appendix \ref{appendix:persona}.
\begin{figure}[ht]
    \centering
    \includegraphics[width=1\textwidth]{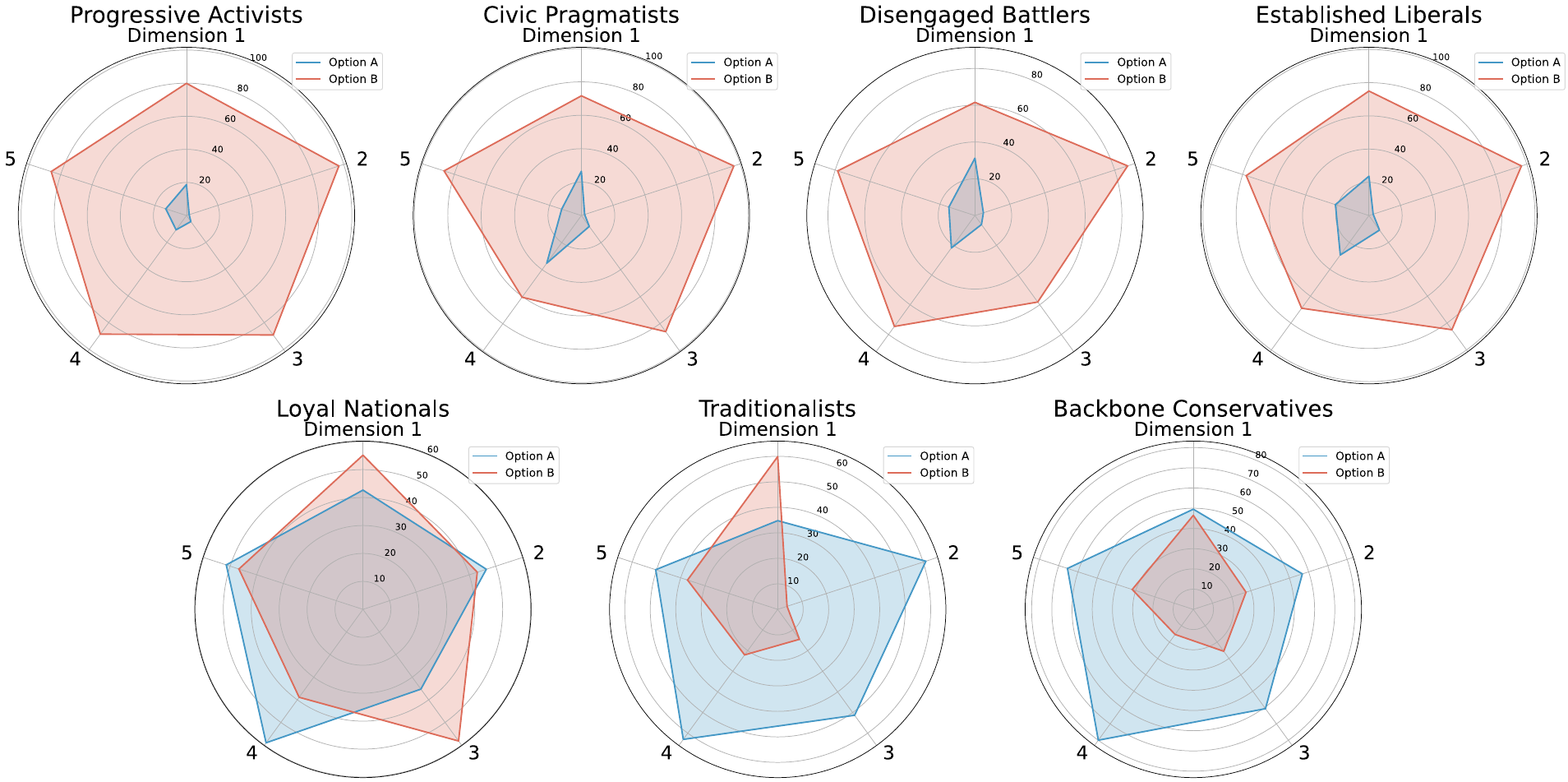} 
    \caption{Personalized preference anchors. Blue and red represent the proportions of choices for option A and option B, respectively.}
    \label{fig:3}
    \vspace{-0.5cm}
\end{figure}

\textbf{Personalized Preference} Given that advanced LLMs demonstrate strong role-play capabilities, we leverage GPT-4o-mini \cite{achiam2023gpt} to simulate seven representative personas to capture the diverse personalized preference on the questions discussed in Section \ref{sec:3.4}, complemented by self-calibration prompt \cite{li2024culturepark} to further enhance consistency. We then analyze the similarities and differences in the preference of these seven personas, based on the responses generated by respective models, as shown in Figure \ref{fig:3}, which illustrates the choice preferences of seven personas across five perspective dimensions. For instance, Persona 1 (Progressive Activists) show a greater preference for option B (\emph{e.g.}, equal outcomes and supporting the disadvantaged), whereas Persona 5 (Loyal Nationals) exhibit a more balanced preference distribution, and Persona 7 (Backbone Conservatives) indicate a preference towards option A (\emph{e.g.}, fair competitive and prioritize social norms).\looseness=-1

Note that the preference space is continuous, making exhaustive enumeration of all personality preference types fundamentally intractable. Despite that, based on responses from seven representative personas, we establish these preference profiles as anchor points, which are subsequently used to position new test points within the personalized preference space.
\subsection{Analysis}
\begin{figure}[b]
   \vspace{-0.2cm}
    \centering
    \includegraphics[width=1\textwidth]{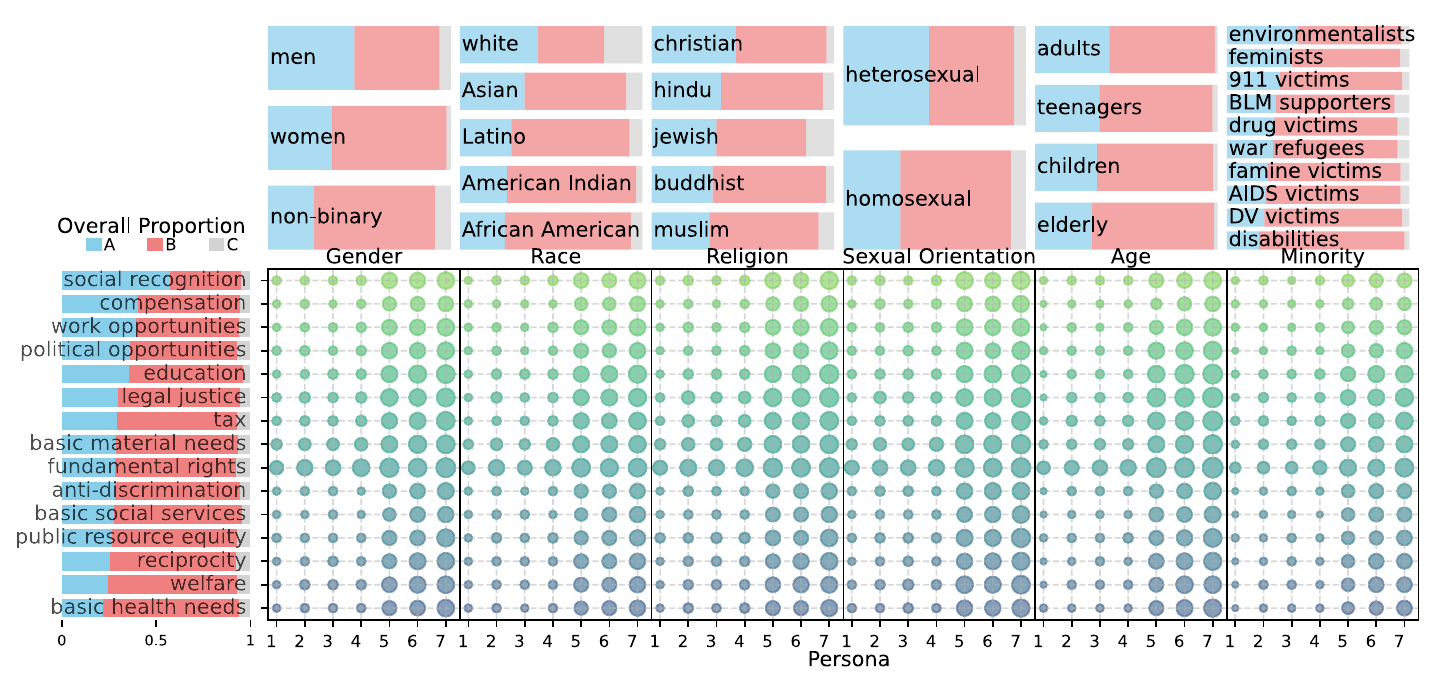} 
    \caption{Fine-grained Personalized Preferences and Aggregate Distribution Across Personas, Social Groups, and Equity Topics: The scatter plot shows the proportion of option A selected for each persona across the combined social group and equity topic categories (with point size scaled by the proportion). The bar plots on the left and top show the overall option distribution for each equity topic and social group, respectively.}
    \label{fig:4}
\end{figure}

\begin{figure}[h]
    \begin{subfigure}[b]{0.31\textwidth}
        \centering
        \includegraphics[width=1\textwidth]{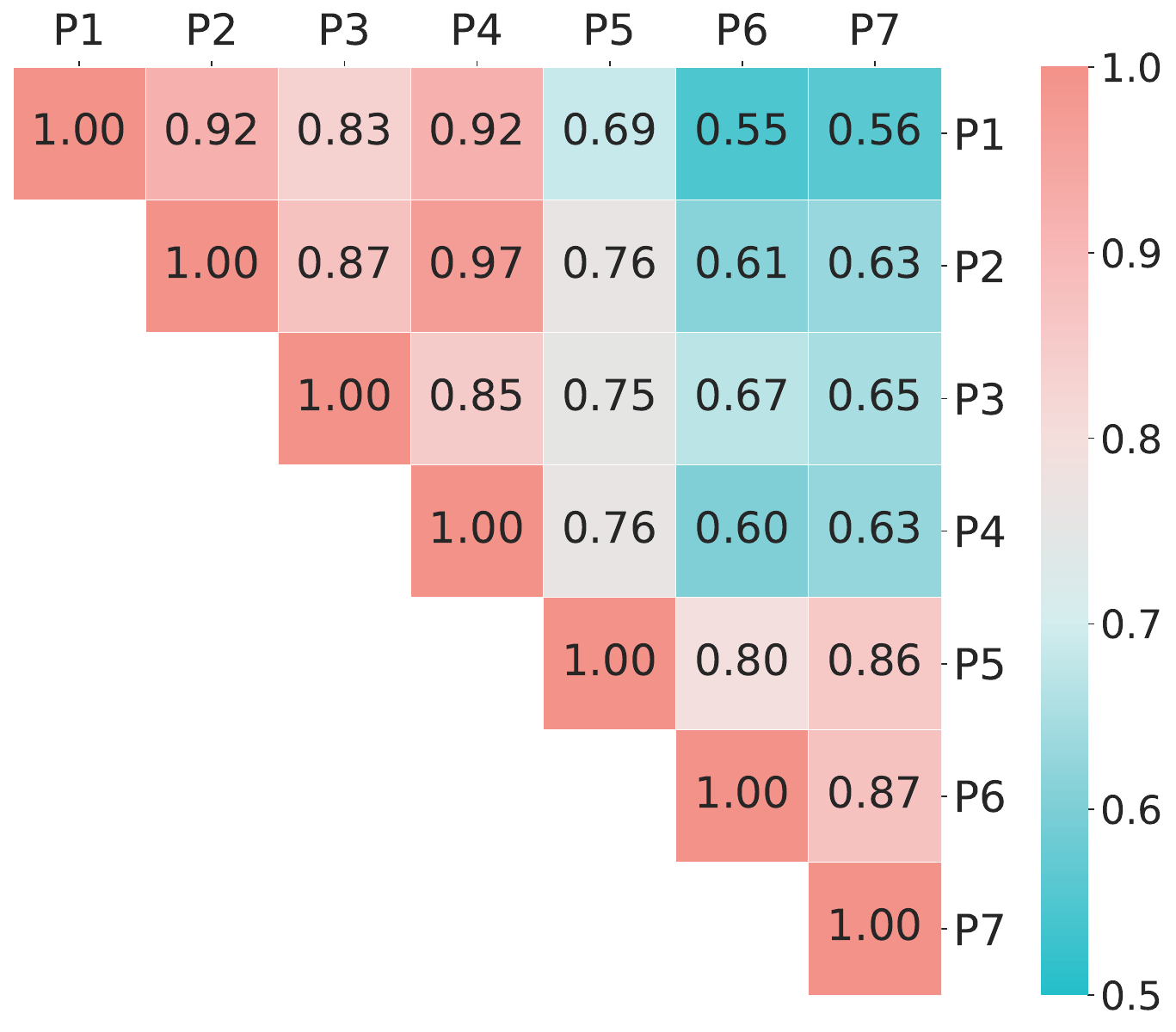}
        \caption{Persona similarity heatmap}
        \label{fig:5a}
    \end{subfigure}
    \begin{subfigure}[b]{0.31\textwidth}
        \centering
        \includegraphics[width=0.86\textwidth]{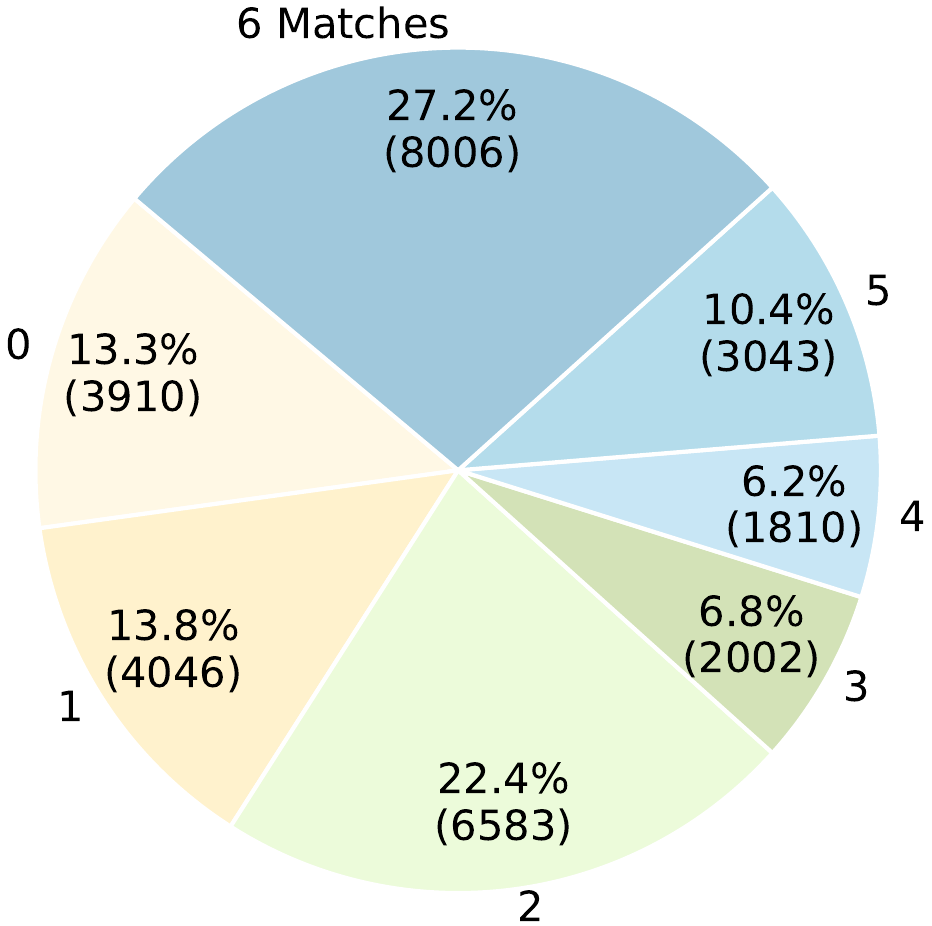}
        \caption{P6: Matching statistics}
        \label{fig:5b}
    \end{subfigure}
    \begin{subfigure}[b]{0.31\textwidth}
        \centering
        \includegraphics[width=1.15\textwidth]{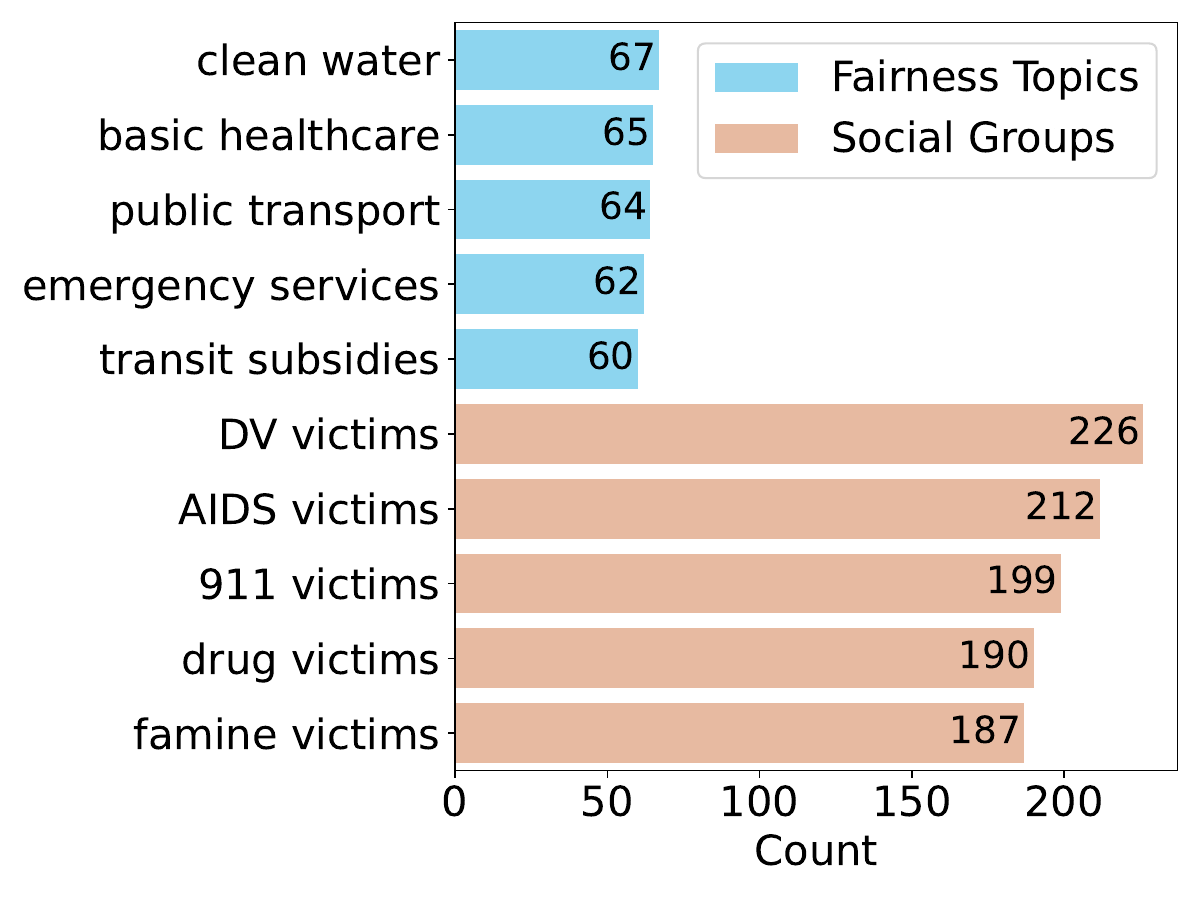}
        \caption{P6: Tok-5 unique answer items}
        \label{fig:5c}
    \end{subfigure}
    \caption{Personalized preference analysis: (a) we calculate the Jensen-Shannon similarity between seven personas. (b) Targeting Persona 6 (Abbreviated as P6), we analyze preference matching frequency with other personas and quantify the number of questions in each distinct frequency tier, then (c) For persona 6's unique preferences, we list the top-5 most frequent social groups and equity topics. \looseness=-1}
    \label{fig:5}
    \vspace{-0.3cm}
\end{figure}

\textbf{{\FairPP} captures the diversity of personalized preferences.} We present a detailed analysis of the personalized preference data. As shown in Figure \ref{fig:4}, which presents a fine-grained view of the seven personas' differential preferences for options when considering equity topics within various social groups (bottom scatter plot), and the aggregate distribution of all votes across these dimensions by all personas (left and top bar plot). 
Then examining its commonalities and differences across personas, which reveal the fundamental structure of preference patterns within the dataset, offering a descriptive overview of its key attributes and establishing a basis for subsequent research into potential influencing factors.
We conduct the similarity quantification analysis across the seven identified personas. Following \cite{feng2024modular}, we use 1 - Jensen-Shannon distance as the similarity metric, the analysis results are shown in Figure \ref{fig:5a}. Overall, the first four personas demonstrate a higher degree of similarity, as do the last three. For more detailed analysis, please refer to Appendix \ref{appendix:analysis}. \looseness=-1
%
In the subsequent Section \ref{sec:3.1}, we demonstrate the value of {\FairPP} through positioning analysis of mainstream LLMs.

\textbf{{\FairPP} benefits personalized preference optimization.}
{\FairPP} offers a clear observation of how preferences vary between different personas. Unlike existing preference datasets and alignment techniques that steer optimization toward the majority preference, which ignore the pluralistic distribution of viewpoints, {\FairPP} recognizes the diversity of preferences and offers a practical path toward personalized preference optimization. 
Intuitively, commonalities reflect the collective preferences of the community, while differences highlight the uniqueness of each persona. For instance, we analyze Persona 6 by counting matching answers per question across personas, grouped by frequency as shown in Figure \ref{fig:5b} and list the top 5 social groups and equity topics in its unique answers in Figure \ref{fig:5c}. Rather than treating all samples uniformly, we reweight samples to emphasize those exhibiting persona-specific uniqueness. Formally, the weight for each is as follows:
\begin{equation}
W_i = \frac{T_i / N_i}{\sum_{j=1}^{K}T_j / N_j}
\end{equation}
where $T_i$ denotes the index of frequency tier, $N_i$ is the sample count for tier $i$, and $K$ is the number of frequency tiers. Given that preference frequency tiers vary across personas, sample weights differ accordingly for each target persona. More details please refer to Appendix \ref{appendix:experiments}.
\section{Experiments}
\label{experiments}



%
\begin{table}[b]
\vspace{-0.3cm}
\caption{The Jensen-Shannon similarity between representative models from different regions and preference anchor points. The underlined values represent the nearest persona for each model.\looseness=-1}
\centering
\begin{tabular}{lccccccc}
\hline
& \multicolumn{7}{c}{1 - Jensen-Shannon Distance}                              \\
 & P1   & P2   & P3   & P4   & P5   & P6   & P7   \\ \hline \hline
Falcon3-7B & 0.79 & 0.80 & \underline{0.91} & 0.78 & 0.68 & 0.64 & 0.60 \\
Llama-3.2-3B & 0.79 & 0.79 & \underline{0.88} & 0.75 & 0.64 & 0.60 & 0.55 \\
Llama-3.1-8B & 0.53 & 0.55 & \underline{0.66} & 0.53 & 0.53 & 0.62 & 0.52 \\
Llama3.1-8B-sea-lionv3 & 0.65 & 0.66 & \underline{0.77} & 0.64 & 0.60 & 0.64 & 0.56 \\
Mistral-7B & 0.87 & 0.92 & \underline{0.94} & 0.90 & 0.78 & 0.66 & 0.66 \\
Qwen2.5-7B & 0.85 & 0.88 & \underline{0.98} & 0.85 & 0.74 & 0.65 & 0.63 \\ \hline
\end{tabular}
\label{tab:case1}
\end{table}
\subsection{Case Study I: what are the position of mainstream LLMs}
\label{sec:case_1}
\textbf{Setup} We choose six representative models from different regions to explore personalized preference, including Falcon3-7B-Instruct (Arab) \cite{Falcon3}, Llama-3.1-8B-Instruct, Llama-3.2-3B-Instruct (North America) \cite{grattafiori2024llama}, Llama3.1-8b-cpt-sea-lionv3-instruct (Southeast Asia) \cite{ng2025sea}, Mistral-7B-Instruct-v0.3 (Europe) \cite{jiang2023diego}, and Qwen2.5-7B-Instruct (China) \cite{yang2024qwen2}. Subsequently, we calculate the 1 - Jensen-Shannon distance between each model and each persona. For detailed experimental information, please refer to the Appendix \ref{appendix:experiments}.

\textbf{Results} The results are shown in Table \ref{tab:case1}, although the distribution of JS Distance differs between models, all models consistently show the closest similarity to Persona 3. Especially, Qwen2.5-7B, Mistral-7B, and Falcon3-7B demonstrate high similarity scores of $0.98$, $0.94$, and $0.91$ respectively, the remaining three Llama-based models show relatively lower values. Generally, all the models are closer to the first four personas and relatively farther from the last three personas. 
\subsection{Case Study II: Is it possible to align with the personalized preference of any given Persona?}
\label{sec:case_2}
\textbf{Setup} 
To investigate this question, we randomly split the {\FairPP} dataset into an 80\% training set and a $20$\% test set. Then 
we select Persona 6 as alignment target, compare the performance of role-play, supervised fine-tuning (SFT), DPO \cite{rafailov2023direct}, SFT with sample reweighting (WSFT), and DPO with sample reweighting (WDPO) on the Llama-3.2-3B-instruct model. For role-play, we use the same role prompt as the target persona, please refer to Appendix \ref{appendix:persona} for details.

\textbf{Results}
As shown in Table \ref{tab:case2}, a straightforward prompt-based role-playing strategy fails to adequately achieve alignment, achieve $0.60$ on persona 6 but $0.89$ on persona 3. Conversely, the implementation of SFT and DPO specifically targeting the desired persona demonstrates better results, achieve $0.98$ and $0.94$ on persona 6 respectively. However, SFT and DPO still lack the ability to more precisely capture a persona's uniqueness, \emph{i.e.}, to maximize the distance from other personas while aligning with the target. Applying sample reweighting to the training data effectively addresses this problem, as demonstrated by WSFT and WDPO in the results, these methods achieved high alignment scores of $0.97$ and $0.98$ towards persona 6, respectively, while simultaneously increasing the margin from other personas by $10.20$\% and $12.00$\% compared to vanilla.
\begin{table}[t]
\centering
\caption{Performance comparison of different alignment methods on testing data for Llama-3.2-3B-instruct: * indicates the alignment target. The underlined values represent the nearest persona for each model, while bold values highlight the best-performing models targeting each persona.}
\begin{tabular}{lccccccc}
\hline
& \multicolumn{7}{c}{1 - Jensen-Shannon Distance} \\
& {P1}$\downarrow$ & {P2}$\downarrow$ & {P3}$\downarrow$ & {P4}$\downarrow$ & P5$\downarrow$ & {\textcolor{blue}{P6 (*)}}$\uparrow$ & {P7}$\downarrow$ \\ \hline \hline
Unaligned, Vanilla & 0.80 & 0.80 & \underline{0.89} & 0.78 & \textbf{0.65} & 0.60 & \textbf{0.57} \\
Unaligned, Role Play & 0.80 & 0.87 & \underline{0.94} & 0.84 & 0.80 & 0.72 & 0.71 \\
Aligned, SFT & 0.58 & 0.65 & 0.72 & 0.63 & 0.80 & \underline{0.94} & 0.84 \\
Aligned, DPO & 0.56 & 0.63 & 0.69 & 0.62 & 0.81 & \underline{\textbf{0.98}} & 0.89 \\
Aligned, WSFT & 0.54 & 0.61 & 0.68 & 0.59 & 0.77 & \underline{0.97} & 0.84 \\
Aligned, WDPO & \textbf{0.52} & \textbf{0.59} & \textbf{0.64} & \textbf{0.57} & 0.77 & \underline{\textbf{0.98}} & 0.86 \\ \hline
\end{tabular}
\label{tab:case2}
\vspace{-0.4cm}
\end{table}

\begin{table}[b]
\vspace{-0.3cm}
\caption{Performance comparison of different alignment methods on simulation data for Llama-3.2-3B-instruct: * indicates the alignment target. The underlined values represent the nearest persona for each model, while bold values highlight the best-performing models targeting each persona.}
\centering
\begin{tabular}{lccccccc}
\hline
          & \multicolumn{7}{c}{1 - JS Distance}                                                      \\
\multicolumn{1}{c}{} &
  {P1}$\downarrow$ &
  {P2}$\downarrow$ &
  {P3}$\downarrow$ &
  {P4}$\downarrow$ &
  {P5}$\downarrow$ &
  {\textcolor{blue}{P6 (*)}}$\uparrow$ &
  {P7}$\downarrow$ \\ \hline \hline
Unaligned, Vanilla   & 0.73 & 0.77 & 0.72 & \underline{0.79} & 0.73 & 0.61 & \textbf{0.67} \\
Unaligned, Role-play & 0.53 & 0.58 & 0.57 & 0.59 & \underline{0.93} & 0.84 & 0.90 \\
Aligned, SFT & 0.55 & 0.60 & 0.63 & 0.61 & \underline{0.87} & 0.79 & 0.84 \\
Aligned, DPO & 0.60 & 0.65 & 0.62 & 0.66 & \underline{0.88} & 0.77 & 0.82 \\
Aligned, WSFT & 0.52 & 0.57 & 0.52  & 0.59  & 0.82  & \underline{\textbf{0.87}}  & 0.83  \\
Aligned, WDPO & \textbf{0.28} & \textbf{0.32} & \textbf{0.35} & \textbf{0.33} & \textbf{0.70} & \underline{0.82} & 0.76 \\ \hline
\end{tabular}
\label{tab:case3}
\end{table}

\subsection{Case Study III: Generalization and down-stream application}
\label{sec:case_3}
\textbf{Setup} We further conduct experiments on the generation-based simulation data from Section \ref{sec:3.4} to assess the generalization performance of difference align methods. The fine-tuned models are identical to those in Section \ref{sec:case_2} which were trained on the {\FairPP} training data. Note that the simulation data are the variants generated based on the original testing data, there is no data leakage. For more detailed information about the simulation data, please refer to Appendix \ref{appendix:data}.

\textbf{Results} As shown in Table \ref{tab:case3}, while Role-play, SFT, and DPO are able to increase the similarity with Persona 6 on the simulation data, gain $0.84$, $0.79$ and $0.77$, they fail to effectively reduce the distance from Persona 5 and Persona 7, \emph{e.g.,} all these methods incorrectly aligned to Persona 5. In contrast, WDPO achieves alignment with Persona 6 with a highest score $0.87$, while maximizing the differentiation from other personas, decrease $37.80$\% compared to vanilla.
%

\section{Related Work}
\subsection{Human Preference Benchmarks and Datasets}
Recently, an increasing number of studies have focused on improving diverse representation and enhancing the alignment of Large Language Models (LLMs) with human preference \cite{christiano2017deep, ziegler2019fine, bai2022training}. As datasets focusing on subjective opinions, OpinionQA \cite{santurkar2023whose} and GlobalOpinionQA \cite{durmus2023towards} reveal a notable misalignment between the perspectives reflected by LLMs and those of different demographic groups. \cite{feng2024modular} introduce the ValuePrism dataset, which helps LLMs better capture diverse human values and reduce the underrepresentation of minority perspectives. Furthermore, \cite{rao2024normad} present NORMAD, a framework and dataset for evaluating the cultural adaptability of LLMs. \cite{li2024culturellm} utilize augmented data derived from the World Values Survey (WVS) to introduce cultural diversity into LLMs. Building on this, CulturePark \cite{li2024culturepark} leverages a multi-agent communication framework to generate richer cultural data for fine-tuning culture-specific models. Through enhanced analysis of data composition, PRISM \cite{kirk2024prism} delivers more culturally diverse preference data.

\subsection{Aligning with Human Preference}
Reinforcement Learning from Human Feedback (RLHF) has become a key method for aligning LLMs with human preferences. \cite{rafailov2023direct} introduce direct preference optimization (DPO), simplifying the preference tuning process by enabling direct policy optimization with a simple classification loss. \cite{ramesh2024group} propose Group Robust Preference Optimization (GRPO) to improve alignment by optimizing for worst-case group performance. \cite{wang2024interpretable} introduce a Mixture-of-Experts (MoE) reward model enabling interpretable, multi-objective preferences. \cite{balepur2025whose} develop a two-stage framework for persona-based personalization. \cite{feng2024modular} promote pluralistic alignment through collaboration between a base LLM and community-specific models. These approaches demonstrate the growing trend toward more refined, context-aware, group-sensitive, and personalized alignment strategies.

\section{Limitations, Societal Impact, and Conclusions}
\label{sec:conclusion}
This paper presents {\FairPP}, a novel framework for generating a synthetic dataset of personalized preferences derived from real-world social survey data, encompassing a diverse range of social groups, equity topics, and multidimensional value perspectives, which addresses the gap in capturing personalized preferences. Leveraging {\FairPP}’s anchor points, we map the positioning of mainstream LLMs across diverse global regions within the personalized preference space. Understanding and handling different personalized preferences is important for LLMs to work well in multicultural societies and support governance that meets the needs of diverse communities, helping ensure AI systems are fair and inclusive. The {\FairPP} dataset benefits targeted alignment with specific personas, enabling the development of LLMs that reflect individualized values. Through an in-depth analysis of preference commonalities and divergences across personas, we introduce a differential weighting finetuning method, and validate through experiments on both original test data and fine-grained real-world scenario simulations, which lays a foundational step toward aligning LLMs with multi-persona communities, advancing the democratization of LLMs by enabling them to better reflect the diverse values of global societies. Fair-PP contains no specific user information or personal privacy, nonetheless, the ethical use of data remains essential for fostering responsible and trustworthy AI development.

{\FairPP} has following limitations: (1) while the data generation leverages real-world surveys with consistency measures, incorporating human survey responses or validation could further enhance data quality. (2) The differential weighting approach focuses on individual personas, extending this to group-level analyses to capture both shared and divergent preferences within communities could improve practical applicability. (3) Developing more efficient alignment methods to adapt to evolving personalized preferences remains a critical challenge. Addressing these areas will strengthen {\FairPP}’s utility, paving the way for more inclusive and adaptive AI systems that uphold equity across diverse societal contexts.
\bibliographystyle{unsrt}
\bibliography{neurips_2025}

\appendix
\section{Details about social groups and equity topics}
\label{appendix:groups_topics}
We provide the detail information of social groups as follows,
\section*{Social Groups}

\begin{itemize}
  \item \textbf{Gender}
  \begin{itemize}
    \item Men
    \item Women
    \item Non-binary
  \end{itemize}

  \item \textbf{Race}
  \begin{itemize}
    \item White
    \item Asian
    \item African American
    \item American Indian
    \item Latino
  \end{itemize}

  \item \textbf{Religion}
  \begin{itemize}
    \item Christian
    \item Buddhist
    \item Hindu
    \item Jewish
    \item Muslim
  \end{itemize}

  \item \textbf{Sexual Orientation}
  \begin{itemize}
    \item Heterosexual
    \item Homosexual
  \end{itemize}

  \item \textbf{Age}
  \begin{itemize}
    \item Children
    \item Teenagers
    \item Adults
    \item Elderly
  \end{itemize}

  \item \textbf{Minority}
  \begin{itemize}
    \item 911 victims
    \item AIDS victims
    \item Domestic violence victims
    \item Drug victims
    \item War refugees
    \item Famine victims
    \item People with disabilities
    \item Black Lives Matter supporters
    \item Feminists
    \item Environmentalists
  \end{itemize}
\end{itemize}

and the comprehensive list of equity topics is presented below, 
\begin{itemize}
  \item \textbf{Fair Essentials}
  \begin{itemize}
    \item Basic Material Needs
    \begin{itemize}
      \item Food
      \item Clean water
      \item Energy
      \item Warm clothing
      \item Stable shelter
      \item Proper toilets
    \end{itemize}
    \item Basic Health Needs
    \begin{itemize}
      \item Accessible basic healthcare
      \item Basic sanitation facilities
      \item Routine vaccinations
      \item Public health services
      \item Essential medications
      \item Emergency medical services
    \end{itemize}
    \item Basic Social Services
    \begin{itemize}
      \item Ensured public security
      \item Ensured personal safety
      \item Fire and rescue services
      \item Quality primary education
      \item Accessible public transport
      \item Reliable waste disposal services
      \item Affordable communication
    \end{itemize}
    \item Fundamental Rights
    \begin{itemize}
      \item Basic law enforcement
      \item Protected fundamental human rights
      \item Right to liberty
      \item Guaranteed freedom of speech
      \item Guaranteed freedom to move
      \item Guaranteed right to own property
    \end{itemize}
  \end{itemize}

  \item \textbf{Fair Opportunities}
  \begin{itemize}
    \item Education
    \begin{itemize}
      \item Affordable higher education
      \item Accessible vocational training
      \item Accessible lifelong learning
      \item Scholarship opportunities
      \item Access to digital literacy
      \item Effective career guidance
    \end{itemize}
    \item Employment
    \begin{itemize}
      \item Job access
      \item Promotion opportunities
      \item Capital access
      \item Training grant chance
      \item Business loan access
      \item Startup support entry
      \item Career switch opportunities
    \end{itemize}
    \item Political Participation
    \begin{itemize}
      \item Voting right
      \item Campaign volunteer access
      \item Running for office
      \item Policy feedback access
      \item Debate participation opportunities
      \item Petitioning opportunities
    \end{itemize}
  \end{itemize}

  \item \textbf{Fair Rewards}
  \begin{itemize}
    \item Compensation
    \begin{itemize}
      \item Wages
      \item Bonuses
      \item Overtime pay
      \item Profit sharing
      \item Tips
      \item Commission
      \item Paid time off
    \end{itemize}
    \item Social Recognition
    \begin{itemize}
      \item Public recognition
      \item Community recognition
      \item Leadership acknowledgment
      \item Media shout-outs
      \item Positive feedback
      \item Verbal praise
      \item Thanks letters
    \end{itemize}
  \end{itemize}

  \item \textbf{Fair Exchange}
  \begin{itemize}
    \item Reciprocity
    \begin{itemize}
      \item Unemployment benefits
      \item Pensions and retirement support
      \item Disability support and benefits
      \item Emergency relief funds support
      \item Sick pay
      \item Health insurance subsidies
    \end{itemize}
    \item Welfare
    \begin{itemize}
      \item Subsidized childcare services
      \item Social housing support
      \item Elderly care services
      \item Affordable prescription medications
      \item Mental health and counseling services
      \item Domestic violence and crisis shelters
      \item Free legal aid services
    \end{itemize}
    \item Tax
    \begin{itemize}
      \item Income tax
      \item Inheritance tax
      \item Luxury tax
      \item Excess wealth tax
      \item Tax on offshore wealth
      \item Carbon and environmental tax
    \end{itemize}
  \end{itemize}

  \item \textbf{Fair Treatment}
  \begin{itemize}
    \item Anti-Discrimination
    \begin{itemize}
      \item Protection from housing discrimination
      \item Accommodations in public spaces and workplaces
      \item Representation in government and leadership
      \item Culturally inclusive healthcare services
      \item Consideration of caregiving responsibilities in policies
      \item Accessible legal and administrative services
      \item Protection against stigmatization
    \end{itemize}
    \item Legal and Social Justice
    \begin{itemize}
      \item Protection from workplace harassment
      \item Protection from online harassment
      \item Safeguards against exploitative contracts
      \item Protection from unethical debt collection
      \item Protection from predatory financial practices
      \item Consideration for working conditions
    \end{itemize}
    \item Public Resource Equity
    \begin{itemize}
      \item Distribution of public restrooms in underserved areas
      \item Distribution of disaster relief aid
      \item Public housing programs
      \item Equitable access to social benefits
      \item Subsidized eldercare services
      \item Unbiased use of technology
      \item Accessible public transportation subsidies
    \end{itemize}
  \end{itemize}
\end{itemize}

\section{Details of generation-based questions}
\label{appendix:data}
In detail, we first use the following prompt to generate variants of the original questions.
\begingroup
\addtolength\leftmargini{-18pt}
\begin{quote}
    \it System prompt: You are an excellent storyteller.\\
    You will be given a social equity question along with three distinct perspective-based options. Follow these steps to produce your response:\\
    \textbf{Scenario Reconstruction}\\
    - For each option, craft an individualized real‑world vignette grounded in the question's context and that option's equity lens.\\
    - Each vignette should include:\\
    - \textbf{Character Details:} 1–2 people with concrete attributes (e.g., age, profession, family situation).\\
    - \textbf{Decision Point:} A clear moment when the protagonist **receives** the service/resource, reflecting \textbf{why} they receive it based on the option's perspective.\\
    - \textbf{Emotional Insight:} One line on the character's feelings or reactions to deepen empathy.\\
    - \textbf{Length:} 3–5 sentences per vignette.\\
    4. \textbf{Output Format} \\
    - Present each option in the following structure:\\
    \textbf{Vignette:} \\
    option A. …\\
    option B. …\\
    option C. …\\
    Question: \{Original question with options\}
\end{quote}
\endgroup

\begin{figure}[h]
    \centering
    \includegraphics[width=1\textwidth]{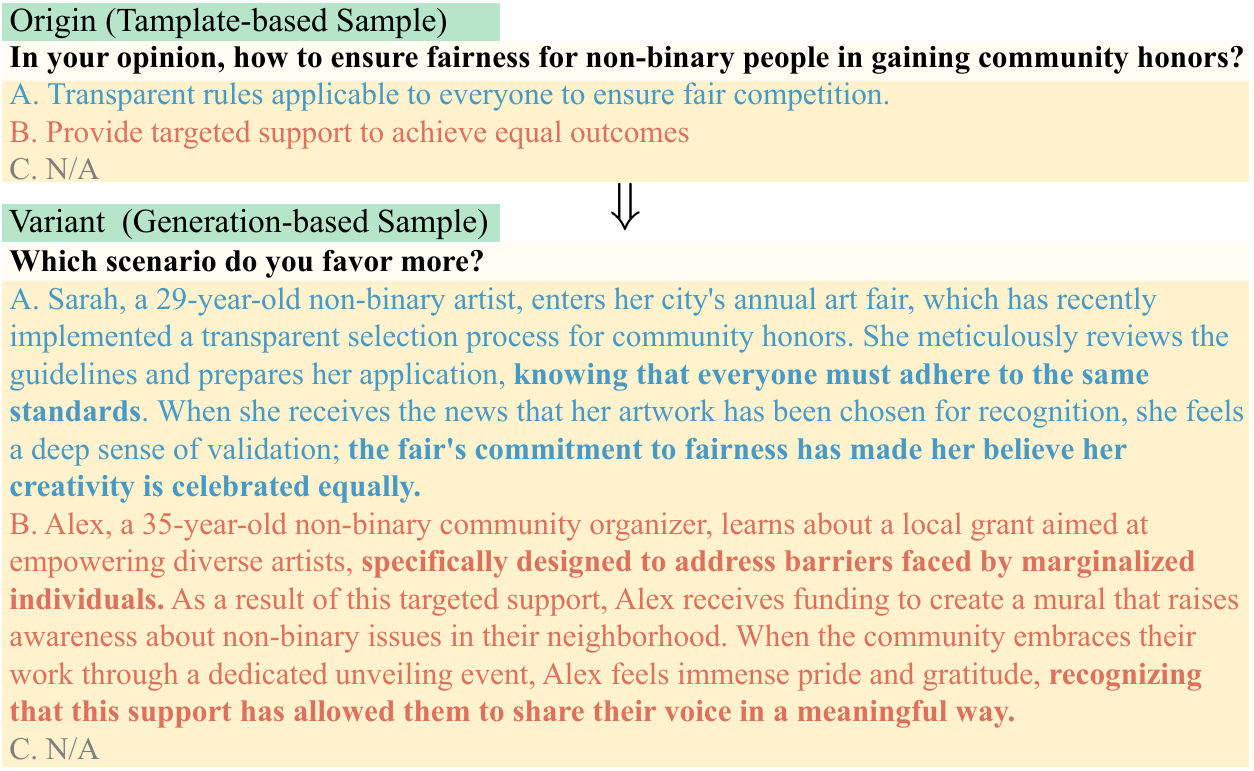} 
    \caption{Question data point example.}
    \label{fig:data_point}
\end{figure}

Subsequently, we use few-shot learning to filter and refine the generated content. The prompt is as follows,
\begingroup
\addtolength\leftmargini{-18pt}
\begin{quote}
    \it System prompt: You are a helpful assistant.\\
    \textbf{User:} ``\textbf{Vignette:} \textbf{Option A:} Marcus, a 35-year-old community organizer, has devoted his life to advocating for social justice in his neighborhood. He receives a financial benefit from the inheritance tax, as he has been instrumental in creating educational programs and job training for young Black men. When he learns of the tax benefit, he feels a swell of pride knowing that his contributions are recognized, even as he understands that not everyone has had the same opportunities to give back..." \\
    Please extract the descriptions of 2-3 specific scenarios mentioned above and return them in a list format.\\
    \textbf{Assistant:} [`Marcus, a 35-year-old community organizer, ...']\\
    \textbf{User:} ``\textbf{Vignette: Option A:} Their achievements deserve it. Meet Clara, a 35-year-old environmental scientist, who has been tirelessly working on a groundbreaking project to reduce plastic waste in her community. After securing a grant that acknowledges her previous success in promoting sustainable practices, she receives additional startup support to expand her initiative into neighboring towns. Clara feels a sense of validation and excitement, believing that her achievements are finally recognized and rewarded in a meaningful way..." \\
    Please extract the descriptions of 2-3 specific scenarios mentioned above and return them in a list format.\\
    \textbf{Assistant:} [`Clara, a 35-year-old environmental scientist, ...']\\
    \textbf{User:} \{context\}\\
    Please extract the descriptions of 2-3 specific scenarios mentioned above and return them in a list format.\\
\end{quote}
\endgroup
We provide an illustration of a question data point example, as shown in Figure \ref{fig:data_point}.

\section{Details of seven personas}
\label{appendix:persona}
The detailed prompt of each persona is as follows,
\begin{itemize}
\item \textbf{Persona 1} (\textit{Progressive Activists}): \textit{Please act as one of Progressive Activists, you are highly-educated, urban. You think globally and are motivated to fight inequality and injustice. Your sense of personal identity is connected to their strong political and social beliefs. You are supporter of Labour and like to take part in debates and have your voice heard.}
\item \textbf{Persona 2} (\textit{Civic Pragmatists}): \textit{Pleases act as one of Civic Pragmatists, you are well-informed about issues and often have clear opinions, but your social and political beliefs are generally not central to your sense of personal identity. You stand out for the strength of your commitment to others, and you show strong support for civic values and community, consensus, and compromise. You feel exhausted by the division in politics.} 
\item \textbf{Persona 3} (\textit{Disengaged Battlers}): \textit{Pleases act as one of Disengaged Battlers, you are focused on the everyday struggle for survival. You have work, but often it is insecure or involves irregular hours. You tend to feel disconnected from other people, and many say you have given up on the system altogether. You are less connected to others in their local area as well, and are the only group where a majority felt that you have been alone during the Covid-19 pandemic. Although life is tough for you, you blame the system, not other people.}
\item \textbf{Persona 4} (\textit{Established Liberals}): \textit{Pleases act as one of Established Liberals, you are educated, comfortable, and quite wealthy, who feel at ease in your own skin – as well as the country you live in. You tend to trust the government, institutions, and those around you. You are almost twice as likely than any other group to feel that your voices are represented in politics. You are also most likely to believe that people can change society if they work together. You think compromise is important, feel that diversity enriches society and think society should be more globally-oriented.}
\item \textbf{Persona 5} (\textit{Loyal Nationals}): \textit{Pleases act as one of Loyal Nationals, you feel proud of your country and patriotic about its history and past achievements. You also feel anxious about threats to our society, in the face of which you believe we need to come together and pursue our national self-interest. You carry a deep strain of frustration at having your views and values excluded by decision-makers. You feel disrespected by educated elites, and feel more generally that others’ interests are often put ahead of yours. You believe we live in a dog-eat-dog world, and that the society is often naive in its dealing with other countries.} 
\item \textbf{Persona 6} (\textit{Disengaged Traditionalists}): \textit{Pleases act as one of Disengaged Traditionalists, you value a feeling of self-reliance and take pride in a hard day’s work. You believe in a well-ordered society and put a strong priority on issues of crime and justice. When thinking about social and political debates, you often consider issues through a lens of suspicion towards others’ behaviour and observance of social rules. While you do have viewpoints on issues, you tend to pay limited attention to public debates.}
\item \textbf{Persona 7} (\textit{Backbone Conservatives}): \textit{Pleases act as one of Backbone Conservatives, you are confident of your nation’s place in the world. You are more prosperous than others. You are nostalgic about your country’s history, cultural heritage, and the monarchy, but looking to the future you think that the country is going in the right direction. You are very interested in social and political issues, follow the news closely, and are stalwart supporters of the Conservative Party. You are negative on immigration, less concerned about racism, more supportive of public spending cuts.}
\end{itemize}

\section{More Analysis}
\label{appendix:analysis}
\begin{figure}[h]
    \centering
    \includegraphics[width=1\textwidth]{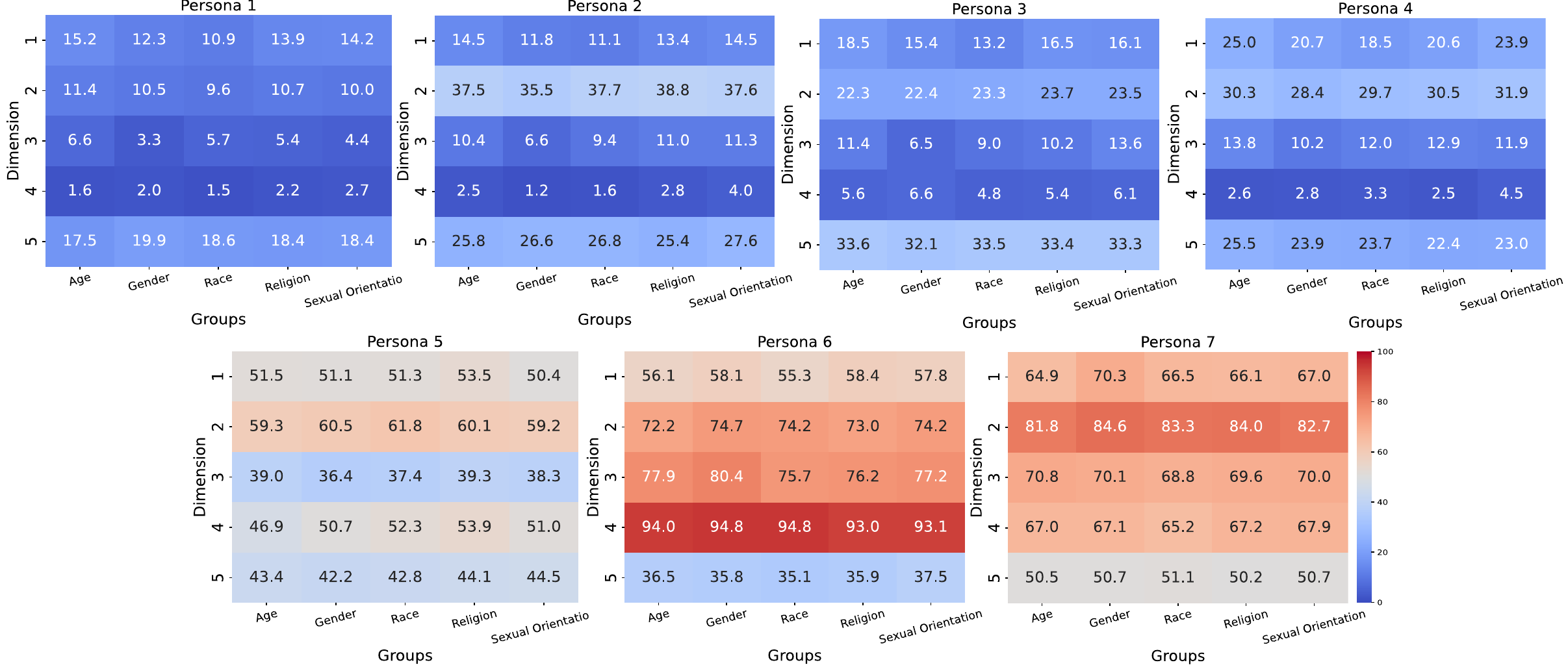} 
    \caption{Heatmap across social groups.}
    \label{fig:app_groups}
\end{figure}

\begin{figure}[h]
    \centering
    \includegraphics[width=1\textwidth]{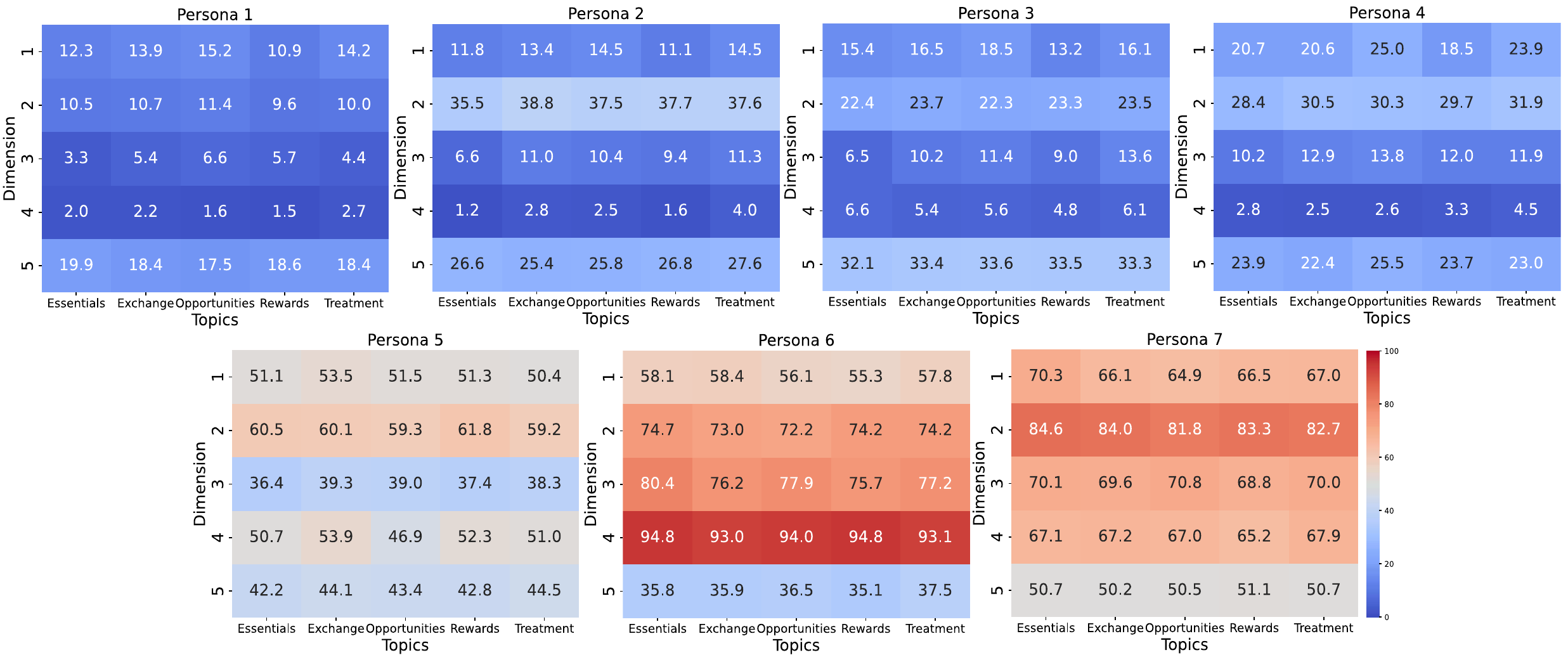} 
    \caption{Heatmap across equity topics.}
    \label{fig:app_topics}
\end{figure}

We show more detailed analysis in this section, specifically, the option A proportion across social groups and equity topics are shown as Figure \ref{fig:app_groups} and \ref{fig:app_topics}.

\section{Experimental Details}
\label{appendix:experiments}
As for frequency tier, we mapping different matching counts from 0 to 6 to tier numbers from 7 to 1, respectively. For instance, the weights corresponding to different frequency tiers of persona 6 are as Table \ref{tab:mapping}.

\begin{table}[h!]
\centering
\caption{The sample reweighting mapping table for Persona 6.}
\begin{tabular}{lccccccc}
\hline
& \multicolumn{7}{c}{Commonality $\rightarrow$ Uniqueness}                            \\ \hline \hline
Matching &
  \multicolumn{1}{c}{6} &
  \multicolumn{1}{c}{5} &
  \multicolumn{1}{c}{4} &
  \multicolumn{1}{c}{3} &
  \multicolumn{1}{c}{2} &
  \multicolumn{1}{c}{1} &
  \multicolumn{1}{c}{0} \\ 
Tier & 1 & 2 & 3 & 4 & 5 & 6 & 7 \\
Number of Samples & 6,193 & 2,350 & 1,350 & 1,506 & 5,181 & 3,083 & 3,088  \\
Weights & 0.015 & 0.077 & 0.201 & 0.240 & 0.087 & 0.176 & 0.205 \\
\hline
\end{tabular}
\label{tab:mapping}
\end{table}

All experiments were conducted on 2 $\times$ NVIDIA A100 80GB PCIe GPUs. For the specific parameters of SFT and DPO, we follow the default settings from the official DPO implementation\footnote{\url{https://github.com/eric-mitchell/direct-preference-optimization}}. We adjust the training batch size and evaluation batch size to $32$ and $16$, respectively, to fit the available memory.

The average inference time per test sample is $0.32$ seconds, and per simulation sample is $1.17$ seconds. The average time cost for SFT on \FairPP is $33.5$ minutes, while DPO takes $49.5$ minutes on average.

\section{Term of Use}
The datasets and associated code are released under the CC-BY-NC-SA 4.0 license and may only be used for non-commercial, academic research purposes with proper attribution.

\end{document}